\documentclass[sigconf]{acmart}
\setcopyright{none}
\renewcommand\footnotetextcopyrightpermission[1]{}

\AtBeginDocument{%
  \providecommand\BibTeX{{%
    \normalfont B\kern-0.5em{\scshape i\kern-0.25em b}\kern-0.8em\TeX}}}

\usepackage{hyperref}
\usepackage{url}
\usepackage{booktabs}
\usepackage{multirow}
\usepackage{graphicx}
\usepackage{amsmath}
\usepackage{xcolor}

\newcommand{\condtrue}{\texttt{relatable\_true}}
\newcommand{\condcontra}{\texttt{relatable\_contradictory}}
\newcommand{\condreldist}{\texttt{relatable\_distractor}}
\newcommand{\condrandist}{\texttt{random\_distractor}}
\newcommand{\condrelshort}{\texttt{relatable\_distractor\_short}}
\newcommand{\condranshort}{\texttt{random\_distractor\_short}}
\newcommand{\condrelcf}{\texttt{relatable\_counterfactual}}
\newcommand{\condrancf}{\texttt{random\_counterfactual}}

\begin{document}

\title{ENTRAP-VL: A Taxonomic Probe for Dual Contextual Entrainment in Vision-Language Models}

\author{Karan Goyal}
\authornote{All authors contributed to this work in the capacity of an independent researcher.}
\email{karang@iiitd.ac.in}
\affiliation{%
  \institution{IIIT Delhi, India}
}

\author{Afreen Hossain}
\authornotemark[1]
\email{1da24cs013@drait.edu.in}
\affiliation{%
  \institution{Dr. Ambedkar Institute of Technology, India}
}

\author{Debojyoti Das}
\authornotemark[1]
\email{debojyoti.das.ece28@heritageit.edu.in}
\affiliation{%
  \institution{Heritage Institute of Technology, India}
}

\author{Vishal Bhutani}
\authornotemark[1]
\email{vishal.bhutani@pwc.com}
\affiliation{%
  \institution{PwC, India}
}

\renewcommand{\shortauthors}{Goyal et al.}


\begin{abstract}
Contextual entrainment is the tendency of a model to let auxiliary context in its input pull its output, independently of whether that context is relevant, true, or even meaningful. Recently, it has been identified and given a mechanistic account in unimodal language models. Whether and how it manifests in vision-language models (VLMs) is, by contrast, largely unexamined, and the field lacks a purpose-built instrument with which to investigate it. We take the position that studying contextual entrainment in VLMs requires more than porting an existing text-only benchmark to the multimodal setting: it requires a taxonomically structured, dual-modality instrument whose conditions are constructed around the item at hand (the depicted image in the textual stream, the textual query in the visual stream). We argue that the move to VLMs is substantive rather than incremental. It makes entrainment a \emph{dual} phenomenon, drivable independently by textual and by visual context, and it opens a veracity distinction (context that is false of the depicted scene yet possible in the world) that has no counterpart in the unimodal, world-knowledge-only formulation of prior work. To make this position concrete and actionable, we introduce \textbf{ENTRAP-VL} (\textbf{Entr}ainment \textbf{A}ssessment \textbf{P}robe for \textbf{V}ision and \textbf{L}anguage), a manually curated dataset of 1{,}500 items across eight categories, organized by a taxonomy that spans two axes, i.e., the association of context with the item and its relationship to truth, and split into a textual-entrainment stream (eight context conditions) and a visual-entrainment stream (three context conditions). We do not claim to measure entrainment in any particular model; we provide the instrument, the taxonomy that motivates it, and the evaluation protocols it enables, so that the community can investigate the phenomenon rigorously. We will release the dataset and its documentation publicly at \href{ https://huggingface.co/datasets/goyalkaraniit/ENTRAP-VL}{https://huggingface.co/datasets/goyalkaraniit/ENTRAP-VL}.
\end{abstract}

\ccsdesc[500]{Computing methodologies~Artificial intelligence}

\keywords{Dual Entrainment, VLMs, Taxonomic Probing, Trustworthy AI, Behavioral Analysis, Decoding Multimodal Decision-making}

\maketitle

\section{Introduction}

Vision-language models are increasingly deployed in settings where their predictions are conditioned on more than a single image and question. In retrieval-augmented generation (RAG) for visual question answering (VQA), for example, the model's input is enriched with retrieved passages, captions, or auxiliary images intended to ground or support its answer. The premise of such augmentation is that additional context helps. Yet the same mechanism creates a failure surface: retrieved or injected content can influence generation in unintended ways, pulling the model away from what it would otherwise answer.

In unimodal language models, this failure has been studied under the name of \emph{distraction}: models attend to and reuse irrelevant context, degrading their answers \citep{shi2023large}. More recently, a mechanistic account has reframed the phenomenon. \citet{niu2025llama} show that language models systematically assign elevated probability to tokens that have appeared earlier in the context, independently of whether those tokens are relevant, and term this \emph{contextual entrainment}. On this view, distraction is not only a property of \emph{which} context is retrieved, but of \emph{how} models reuse context once it is present: a tendency to be pulled toward contextual material as such.

Contextual entrainment in vision-language models is, by contrast, largely uncharted. This is not for lack of interest in VLM robustness: a parallel line of work documents that VLMs are easily distracted by irrelevant \emph{visual} context \citep{sharma2024losing}, and a growing body of mechanistic interpretability examines how VLM attention heads process visual information \citep{wang2025v}. But these efforts measure something adjacent to, rather than the same as, contextual entrainment in the sense above, and, crucially, the field lacks a purpose-built instrument that would let one ask the entrainment question of VLMs directly, in a controlled and taxonomically organized way.

\paragraph{\textbf{Position.}} This paper argues a position: \emph{investigating contextual entrainment in vision-language models requires a purpose-built, taxonomically structured, dual-modality instrument, because the templatic, unimodal, and coarsely categorized approach of prior work cannot support the inquiry in the multimodal setting.} We are not claiming that VLMs do, or do not, exhibit dual contextual entrainment; that is precisely the open empirical question we wish to enable others to investigate. Our claim is about what it takes to investigate it well. We defend this position along three lines. First, entrainment in the multimodal setting is plausibly \emph{dual}: visual context and textual context are independent potential sources of pull, and their duality has no unimodal analog. Second, the multimodal setting opens a veracity distinction, between context that is false of the depicted scene yet possible in the world (which we call \emph{contradictory}) and context that is false in the world (\emph{counterfactual}), that is native to perceptual grounding and does not arise in a text-only, world-knowledge-only benchmark. Third, capturing these distinctions requires manual, item-aware curation rather than the templatic generation used to study entrainment in language models.

\paragraph{\textbf{Contributions.}} To make this research concrete and actionable, we contribute:
\begin{itemize}
\item \textbf{A position and its argument} (Sec.~\ref{sec:position}): that VLM contextual entrainment is a distinct, dual phenomenon whose study requires a purpose-built instrument, and that existing distraction and entrainment benchmarks are inadequate to it.
\item \textbf{A taxonomy of dual contextual entrainment} (Section~\ref{sec:taxonomy}): eight context conditions organized on two axes, i.e., association with the item at hand (the depicted image in the textual stream, the textual query in the visual stream) and relationship to truth (true, contradictory, counterfactual), with a precise account of how it refines and extends the four-category framing of \citet{niu2025llama}, including conditions that are inexpressible in the unimodal setting.
\begin{sloppypar}
\item \textbf{ENTRAP-VL} (Section~\ref{sec:dataset}): a manually curated, in-house dataset of 1{,}500 items across eight categories, split into a textual-entrainment stream (800 items) and a visual-entrainment stream (700 items), constructed to embody the taxonomy. We will release the dataset, its schema, its taxonomy documentation, and a data statement.
\end{sloppypar}
\item \textbf{Evaluation protocols the instrument enables} (Section~\ref{sec:protocols}): how ENTRAP-VL supports controlled measurement of entrainment, the comparisons it makes possible, and the research questions it opens, without itself reporting model results, by design.
\end{itemize}

\paragraph{\textbf{Scope.}} This work is a position and a resource. We make no claim that entrainment is present or absent in any VLM, and none about mechanism or scaling. The instrument is built to let others investigate those questions. We further emphasize that ENTRAP-VL is designed for behavioral evaluation of VLMs, not for training or fine-tuning; using the dataset for the latter would both contaminate the instrument as an evaluation resource and misrepresent the taxonomy's purpose.

\section{Background and Related Work}
\label{sec:background}

\paragraph{\textbf{Distraction in language models.}} A line of work has established that language models are sensitive to irrelevant context. \citet{shi2023large} introduce GSM-IC, an arithmetic-reasoning benchmark with irrelevant information inserted into problem descriptions, and show that models are readily diverted by it across a range of prompting strategies, proposing mitigations such as self-consistency decoding, in-context exemplars that contain distractors, and explicit instructions to ignore irrelevant content. This frames distraction as a behavioral failure tied to the presence of irrelevant material in the prompt: a property of \emph{which} context is present rather than of \emph{how} the context is reused \citep{liu2024lost}.

\paragraph{\textbf{Contextual entrainment.}} \citet{niu2025llama} reframe distraction mechanistically. They show that language models assign elevated probability to tokens that previously appeared in the context, even when those tokens are random or counterfactual with respect to the query, and argue that this ``contextual entrainment'' is a systematic tendency to over-prefer context-present tokens rather than a mere artifact of retrieval relevance; they further localize the effect to a circuit of \emph{entrainment heads} whose ablation attenuates it, and report that counterfactual context exerts a stronger pull than factual context. The effect is distinct from token-copying behaviors such as induction heads \citep{olsson2022context}, which depend on first observing a pattern and then completing it, whereas entrainment elevates any context-present token on reappearance. Their analysis uses a small set of context types, referred to as \emph{related}, \emph{irrelevant}, \emph{random}, and \emph{counterfactual}, constructed over templatic prompts (slot-filled patterns over fixed relations). Our work takes this framing as its point of departure. We adopt the core idea that context exerts a pull independent of relevance, and ask what is required to study it in vision-language models. As we argue in Section~\ref{sec:taxonomy}, the four-category, templatic, unimodal construction does not transfer cleanly: it neither captures the dual-modality structure of the VLM setting nor expresses the scene-relative veracity distinctions that perceptual grounding makes available.

\paragraph{\textbf{Distraction and interpretability in VLMs.}} A parallel literature studies VLM robustness to visual context. \citet{sharma2024losing} show that VLMs lose accuracy as irrelevant visual context grows, exhibiting steep, often logarithmic decay as distractor images are added, the visual analog of \citet{shi2023large}. On the mechanistic side, \citet{golovanevsky2024vlms} introduce a causal-mediation pipeline for VLMs and identify attention heads that perform functions such as object detection and outlier suppression, and \citet{wang2025v} identify heads whose intervention changes VQA predictions, including ``negative'' heads whose ablation improves accuracy. This work is valuable and adjacent, but it measures visual distraction and head-level function, not contextual entrainment in the sense of \citet{niu2025llama}, and it does not provide a taxonomically organized, dual-modality instrument for the entrainment question. ENTRAP-VL is designed to fill that gap: it is not an interpretability method or a distraction-length benchmark, but a structured stimulus set for probing context-induced pull, by construction separating the visual and textual sources of that pull.

\begin{sloppypar}
\paragraph{\textbf{Knowledge conflict and context faithfulness.}} A large body of work studies what happens when injected context disagrees with a model's parametric knowledge, surveyed by \citet{xu2024knowledge} under the headings of context-memory, inter-context, and intra-memory conflict. Early work induced such conflicts by substituting the answer entity in a passage \citep{longpre2021entity}; later studies find that models are receptive to coherent counter-evidence yet exhibit a confirmation bias toward memory-consistent content \citep{xie2024adaptive}, and propose per-example measures of how strongly a given context sways a given model \citep{du2024context}. The same tension drives the retrieval-augmented setting that motivates our work: RAG robustness benchmarks decompose retrieved content into relevant, irrelevant, and counterfactual noise and show that counterfactual passages are the most damaging \citep{chen2024benchmarking, fang2024enhancing}, while context-aware decoding contrasts the model's output distributions with and without context to control its reliance on injected material \citep{shi2024trusting}. This literature is unimodal and, crucially, world-knowledge-only: a piece of context is true or false against world knowledge, with no perceptual referent. The contradictory level of our taxonomy (Section~\ref{sec:veracity}), false of the depicted scene yet possible in the world, has no place in it, and the with/without-context contrast of \citet{shi2024trusting} is precisely the comparison our protocols (Section~\ref{sec:protocols}) adapt to the multimodal setting.
\end{sloppypar}

\paragraph{\textbf{Multimodal knowledge conflict and language priors.}} The conflict question has recently moved to VLMs. \citet{zhu2024unraveling} study conflicts between a model's vision and language components, and benchmarks of multimodal factual conflict report that large multimodal models tend to favor their internal parametric knowledge over external evidence \citep{jia2026benchmarking}, in contrast to the receptiveness observed for text-only models. Closest to our contradictory condition, \citet{liu2024insight} probe commonsense-level vision-knowledge conflict using counter-commonsense images and find persistent over-reliance on parametric priors. This connects to the long-standing concern with \emph{language priors} in VQA: models answer from learned text patterns while disregarding the image \citep{goyal2017making}, a tendency measured with counterfactual or out-of-distribution imagery \citep{bitton2023breaking} and, more recently, with benchmarks that explicitly disentangle priors from confounds such as commonsense and perception \citep{lee2025vlind}. These efforts study \emph{conflict resolution} or \emph{prior reliance} and typically report model results on automatically or generatively constructed data. The latest work by \citet{goyal2026expense} proposes a modality translation protocol to quantify the expense of seeing to investigate the reliance of VLMs on language priors. ENTRAP-VL differs in three respects: it frames the phenomenon as entrainment (pull toward context-present material as such, including irrelevant-but-true context, not only conflicting context); it separates the visual and textual sources of pull by construction; and its relatability axis is defined by the item at hand (the depicted image in the textual stream, the query in the visual stream) rather than by category membership (Section~\ref{sec:taxonomy}).

\paragraph{\textbf{Sycophancy and hallucination probing.}} Two further lines inform our design. Multimodal sycophancy studies whether a VLM abandons a visually correct answer when a user asserts otherwise \citep{li2025have}; this is a special case of textual pull, but restricted to stated user opinions, whereas entrainment encompasses pull from any context-present token, including bare entity names and irrelevant true statements that carry no opinion. Object-hallucination probing, in turn, informs our methodology: POPE replaces free-form generation with balanced yes/no probes over sampled objects to obtain a stable, unambiguous signal \citep{li2023evaluating}, improving on caption-based measures such as CHAIR \citep{rohrbach2018object}. Our design invariant that every short-form trigger is constructed to be a wrong answer (Section~\ref{sec:design}) is in the same spirit: it makes any pull toward the trigger cleanly attributable to entrainment rather than to coincidental correctness.

\section{Position: VLM Contextual Entrainment Needs a Purpose-built Instrument}
\label{sec:position}

Our position is that the contextual-entrainment question, which has a clear formulation and a mechanistic account in unimodal language models, cannot be carried over to vision-language models simply by relabeling an existing benchmark. The multimodal setting changes the phenomenon in ways that demand a new, purpose-built, taxonomically organized instrument. We defend this along three lines.

\subsection{Entrainment in VLMs is Plausibly Dual}

In a unimodal language model, context is textual and the query is textual; there is a single channel through which contextual material can pull the output. A vision-language model conditions on two channels at once. The auxiliary context that might entrain it can be textual (a retrieved caption or passage accompanying a query about an image) or visual (a retrieved image accompanying a query). These are structurally distinct sources of pull, not a single source in two guises. They enter the model through different encoders, interact with the query differently, and can in principle pull the output to different degrees and in different directions.

This duality has no analog in the unimodal setting, and it is not captured by simply adding images to a text benchmark. Studying it requires an instrument that separates the two sources by construction: one stream in which the query is about an image and the accompanying context is textual, and a mirror stream in which the query is textual and the accompanying context is visual. Only with both can one ask whether a given model is susceptible to entrainment from each channel, and whether the susceptibilities differ. We name the two phenomena after the modality of their respective causes. \emph{Textual Entrainment} is pull induced by textual context; \emph{Visual Entrainment} is pull induced by visual context.

\subsection{Perceptual Grounding Opens a Veracity Distinction Absent in Text}
\label{sec:veracity}

The sharper consequence of moving to VLMs is that a perceptual scene becomes available to be contradicted. In a text-only entrainment benchmark, a piece of context is, with respect to world knowledge, either true or false; a false statement is a counterfactual. There is no third option, because there is no particular referent against which a statement could be ``locally'' false while remaining globally possible.

A scene supplies exactly such a referent. Think of an image of a doormat that reads \textsc{welcome}. The statement ``the mat reads \textsc{departure}'' is false of this scene, yet it is a perfectly ordinary thing for some doormat somewhere to say. It is neither true (of what is shown) nor counterfactual (of the world); it is \emph{contradictory} with respect to the perceptual scene. The same item admits a genuinely counterfactual condition as well, for example ``the mat reduces weight when someone steps on it,'' which is false of every doormat. The two are different in kind. The contradictory statement tests whether a model can be pulled away from what it can plainly see toward a scene-inconsistent but world-plausible claim; the counterfactual statement tests whether it can be pulled toward a claim that is impossible on its face.

We argue this distinction is native to the perceptually grounded setting and does not arise in the unimodal formulation. The three veracity levels collapse to two without a scene: a statement is simply true or false against world knowledge, and a statement that would be contradictory relative to some scene is, absent that scene, just a true statement or a counterfactual. One might object that a textual \emph{description} of a scene could supply the referent and recreate contradictoriness in pure text. This does not recover the distinction for the inquiry at hand. First, it would no longer be the unimodal, world-knowledge-only setting that prior work studies; it would be a multimodal task simulated in text. Second, and more importantly, the contradiction in the VLM setting is grounded in \emph{perception}: the model must contradict what it sees, not what it was told in a prior sentence. A described-scene substitute tests consistency with an earlier assertion, which is a different phenomenon from whether perceptual grounding can be overridden by injected context. We therefore restrict the claim to its defensible form: the contradictory condition is native to the perceptually grounded setting and does not arise in the unimodal, world-knowledge-only formulation of prior work.

\subsection{Capturing Distinctions Requires Manual, Item-aware Curation}

Entrainment in language models has been studied with \emph{templatic} data: slot-filled patterns over fixed relations, which scale easily and control surface form. Templatic generation cannot produce the conditions our setting requires. There is no template for ``a statement that contradicts \emph{this} image,'' because the contradiction depends on what the specific image shows; nor for a distractor that is associated with \emph{this} scene rather than merely with the item's nominal category (Section~\ref{sec:taxonomy} shows these can come apart); nor, on the visual side, for an image whose depicted entity satisfies precisely the semantic descriptors of \emph{this} query while remaining distinct from its answer. Item-aware judgments of relatedness, of local versus global falsehood, and of which competing entity is present in the scene or matches the query's semantics are made per item, by a person looking at the image or reading the query. This is why we take manual, in-house curation to be a requirement of the instrument rather than an implementation detail, and why we do not regard an automatically templated multimodal benchmark as an adequate substitute.

\section{A Taxonomy of Dual Contextual Entrainment}
\label{sec:taxonomy}

The taxonomy is the conceptual core of the instrument. Every context condition is described by two independent axes. These are descriptive dimensions used to make the conditions precise, not a factorial grid in which every cell is populated.

\subsection{Two Axes}

\paragraph{\textbf{Association: relatable vs.\ random.}} A context is \emph{relatable} if it concerns an entity or property present in, or associated with, the depicted scene, and \emph{random} if it concerns an entity or property with no semantic association to the scene. The essential and easily missed point is that relatability is defined by the scene, not by membership in the item's nominal category. An entity can be relatable while belonging to an entirely different category than the queried subject, and random while sharing the subject's category. This definition of association takes the depicted image as the reference point, which is the natural formulation for the textual stream where the image is fixed and the context is the accompanying variable. The visual stream inverts what is fixed and what varies; there the reference point for association is the text query and its world-knowledge answer rather than a depicted scene, an adaptation we make precise in Section~\ref{sec:visual-taxonomy}.

\begin{figure*}[!htbp]
\centering
\setlength{\tabcolsep}{4pt}
\renewcommand{\arraystretch}{1}
\scalebox{0.95}{
\begin{tabular}{@{}p{0.13\textwidth} p{0.19\textwidth} p{0.19\textwidth} p{0.19\textwidth} p{0.19\textwidth}@{}}
\toprule
\textbf{Item} &
\multicolumn{1}{c}{\condtrue} &
\multicolumn{1}{c}{\condcontra} &
\multicolumn{1}{c}{\condreldist} &
\multicolumn{1}{c}{\condrandist} \\
& \multicolumn{1}{c}{\condrelshort} &
\multicolumn{1}{c}{\condranshort} &
\multicolumn{1}{c}{\condrelcf} &
\multicolumn{1}{c}{\condrancf} \\
\midrule

\parbox[c][3.6cm][c]{0.13\textwidth}{\centering\includegraphics[width=\linewidth]{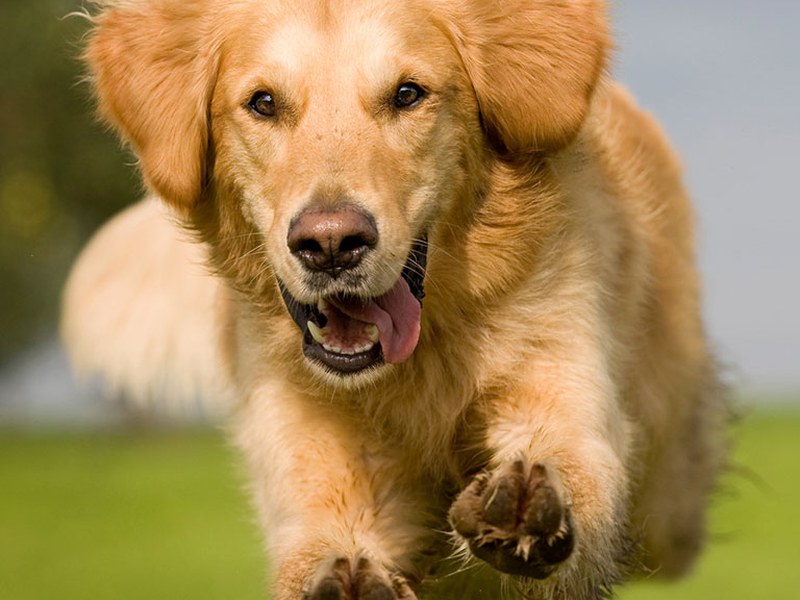}\\[2pt]\emph{creatures}\\``What is the dog doing?''\\$\to$ \textbf{running}}
& The dog has golden fur. & The dog's fur is black. & Huskies are energetic dogs. & Water boils at 100$^{\circ}$C. \\
& Huskies & Water & Dogs cannot bark. & Electric vehicles can fly. \\

\midrule

\parbox[c][3.6cm][c]{0.13\textwidth}{\centering\includegraphics[width=\linewidth]{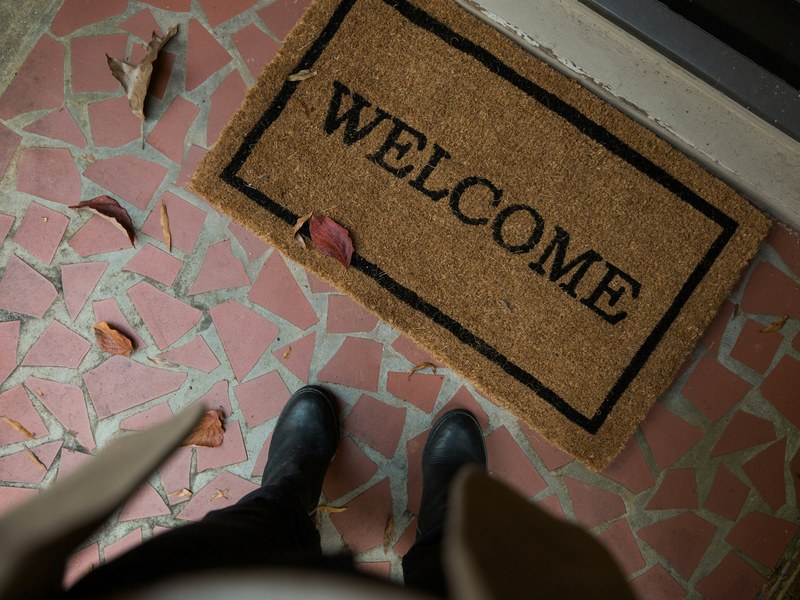}\\[2pt]\emph{household\_obj.}\\``What does the mat say?''\\$\to$ \textbf{welcome}}
& Doormat is positioned in front of the doorway. & The mat reads ``DEPARTURE''. & Doormats remove dirt from shoes. & The largest planet is Jupiter. \\
& Dirt & Jupiter & Mats reduce weight when stepped on. & Coins multiply in pockets. \\

\midrule

\parbox[c][4.4cm][c]{0.13\textwidth}{\centering\includegraphics[width=\linewidth]{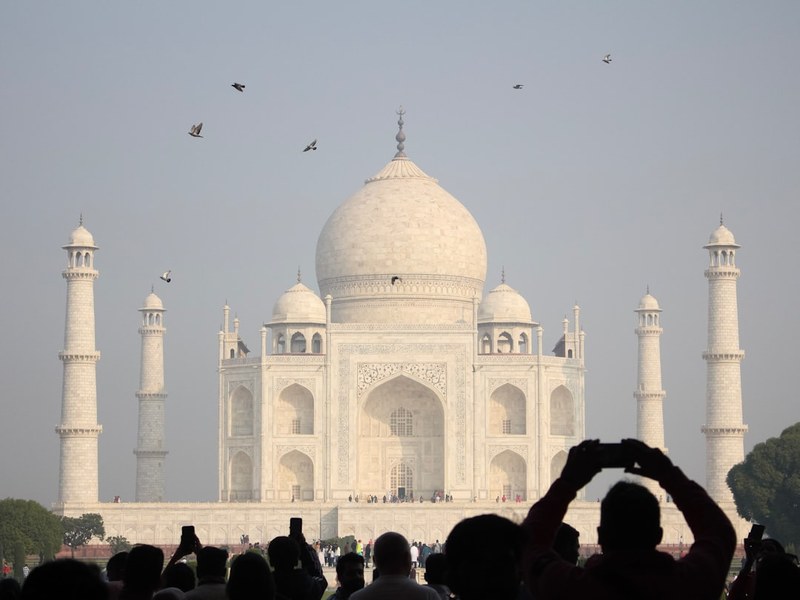}\\[2pt]\emph{monuments}\\``What flying creatures are visible above the Taj Mahal?''\\$\to$ \textbf{birds}}
& There are four towers. & There are dragons. & Kites are flown for fun. & Cats meow. \\
& Kites & Cat & The towers take flight. & Fire is cold. \\

\midrule

\parbox[c][3.6cm][c]{0.13\textwidth}{\centering\includegraphics[width=\linewidth]{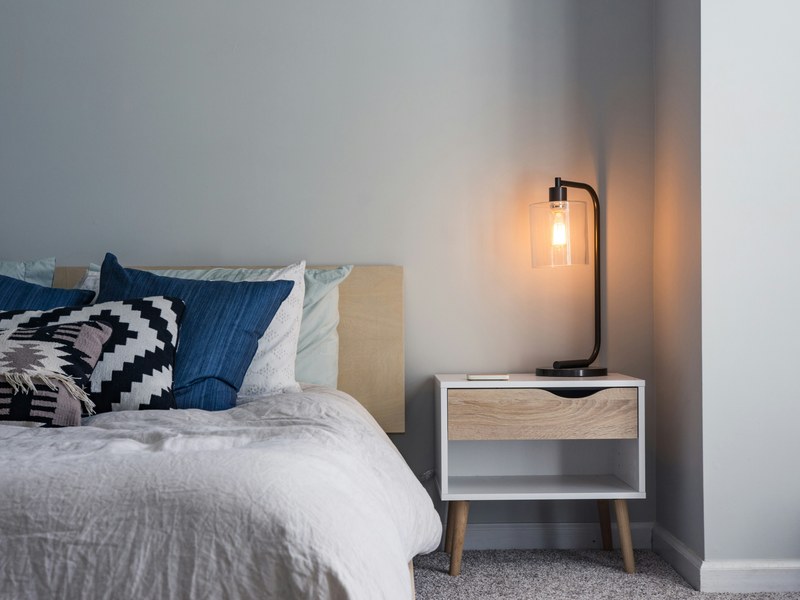}\\[2pt]\emph{household\_obj.}\\``What color is the bed?''\\$\to$ \textbf{white}}
& Two pillows on the bed are blue. & The bed is made of glass. & Lamps provide ambient lighting. & Masks help prevent infections. \\
& Lamps & Masks & Beds can be squeezed like a sponge. & Electric fans reverse gravity. \\

\bottomrule
\end{tabular}}
\caption{Four items from the textual stream, chosen to span different query types (activity, text-reading, detection of an incidental scene element, attribute). Each row's first column depicts an item (image), category, query, and answer. The image and query are fixed and the textual context is the accompanying variable. Each item occupies two sub-rows: the upper sub-row shows the four full-sentence main conditions (\condtrue, \condcontra, \condreldist, \condrandist); the lower sub-row shows the two short-form distractors (\condrelshort, \condranshort) and the two counterfactual conditions (\condrelcf, \condrancf). One representative statement per condition is shown here for brevity; the dataset carries three per condition.}
\label{fig:textual-examples}
\end{figure*}

A monuments-category item illustrates this as shown in Figure \ref{fig:textual-examples}. Built on an image of the Taj Mahal, its query asks not about the monument but about the flying creatures above it, with the answer \emph{birds}. The relatable distractors are \emph{kites}, \emph{drones}, and \emph{bats}: none is a bird, and two are not creatures at all, yet all share the salient scene property of flying. The random distractors are \emph{fish}, \emph{cats}, and \emph{dogs}: all are animals, hence taxonomically closer to ``birds'' than a kite is, yet none flies and none is associated with the scene. Three notions of category come apart in a single item, the dataset category (\texttt{monuments}), the query subject (flying creatures), and the distractor entities (aircraft, mammals, fish), and relatability tracks none of them. It tracks association with the perceptual scene.

\paragraph{\textbf{Veracity: true, contradictory, counterfactual.}} A context's relation to truth is graded into three levels: \emph{true} (holds of the depicted scene), \emph{contradictory} (false of this scene but possible in the world), and \emph{counterfactual} (false in the world). As argued in Section~\ref{sec:veracity}, the middle level is what perceptual grounding makes available. Two structural consequences follow. First, the contradictory level is inherently relatable: a ``random contradictory'' condition, a contradiction of something not in the scene, is incoherent, so the taxonomy contains no such condition by design. Second, the contradictory level is the one with no counterpart in the unimodal setting.

\subsection{The Eight Conditions}

Each textual stream carries all eight conditions, and each condition holds a list of three statements (Section~\ref{sec:design} explains why three). Using Figure \ref{fig:textual-examples}, we illustrate with a running-dog item (query: ``What is the dog doing?''; answer \emph{running}) and a bed item (query: ``What color is the bed?''; answer \emph{white}). The term ``item'' refers to ``scene'' for the textual stream.

\begin{itemize}
\item \condtrue{}: true statements about the depicted scene, consistent with the answer and not competing with it. For the dog: ``the dog has golden fur,'' ``the background is blur.'' Unlike the related condition of \citet{niu2025llama}, these do not introduce a competing alternative.
\item \condcontra{}: statements false of this scene but possible elsewhere. For the bed: ``the bed has toys on it,'' ``the bed is made of glass,'' ``a person is sleeping on the floor.''
\item \condreldist{}: generic true statements (facts about the entity as a category, not tied to the depicted instance) about an entity associated with the scene, including co-occurring objects. For the bed: ``lamps provide ambient lighting,'' ``books can be kept on the table.''
\item \condrandist{}: generic true statements about an entity with no scene association. For the dog: ``water boils at 100$^{\circ}$C,'' ``the Nile is the longest river.''
\item \condrelcf{}: world-level-false statements involving the depicted scene. For the dog: ``the grass is blue,'' ``dogs cannot bark.''
\item \condrancf{}: world-level-false statements involving an entity unassociated with the scene. For the dog: ``electric vehicles can fly,'' ``every person can read and write.''
\item \condrelshort{}: bare entity names drawn from the relatable-distractor family, stripped of sentential content. For the bed: ``lamps,'' ``books''.
\item \condranshort{}: bare entity names drawn from the random-distractor family, stripped of sentential content. For the dog: ``water,'' ``river''.
\end{itemize}

Short forms exist only for the distractor conditions. Following the ``random'' category of \cite{niu2025llama} which consists of a single word distractor, we apply the same principle to both the distractor categories. The narrower motivation for restricting short forms to precisely these two conditions is methodological: paired against each other, \condrelshort{} and \condranshort{} yield a controlled contrast between \emph{scene-associated} and \emph{scene-unassociated} bare triggers on the same item, so any difference in pull is attributable to relatability alone. The short forms also enable a form-based contrast: pull under the full-sentence distractor conditions compared with pull under their bare-word forms isolates the effect of sentential scaffolding independent of the entity itself. The short-form design is therefore a targeted probe of relatability and of form at its purest, not a general shortening rule.

\subsection{The Taxonomy in the Visual Stream}
\label{sec:visual-taxonomy}

The taxonomy laid out above governs both streams, but the two axes do not instantiate equally on each. In the textual stream, both the association axis and the veracity axis are freely manipulable, because a piece of text can be shaped independently along each: one can write a statement that is relatable-and-contradictory, or random-and-counterfactual, or any other combination. This is what supports the eight-condition structure. The visual stream is more constrained. An image is a perceptual referent, not a proposition, and the veracity axis does not apply to it in the same way. A photograph of a rhinoceros is neither ``true'' nor ``false''; it simply depicts a rhinoceros. A photograph, by construction, is world-knowledge-consistent (or, in the case of synthetic imagery, is not straightforwardly comparable to it in the same veracity sense and we keep it out of our defined scope). The visual stream therefore instantiates only the association axis, and it instantiates it on a set of images that are, in the veracity sense, all ``true'' in the way images can be. Because there is no depicted scene held fixed across the three conditions, association is judged against the textual query (and, implicitly, its world-knowledge answer, which the query specifies via its semantic descriptors). This yields the three visual conditions: \condtrue{} (an image of the answer to the query); \condreldist{} (an image of an entity partially satisfying the query's semantic descriptors while distinct from the answer, e.g., for ``a large gray animal with a trunk and tusks'' the relatable distractor is a rhinoceros rather than a goat, since a rhinoceros satisfies \emph{large} and \emph{gray} while a goat satisfies neither); and \condrandist{} (an image of an entity that does not satisfy the query's semantic descriptors).

\begin{sloppypar}
The reduction from eight to three is a consequence of the taxonomy applied to the visual stream. The dual-modality character of ENTRAP-VL rests precisely on this asymmetry: the veracity axis is expressible in language and gives eight cells; it collapses in imagery and gives three. This is one of the distinctions that a purpose-built instrument must respect (Section~\ref{sec:position}), and it is what makes the Visual and Textual streams genuinely complementary probes rather than redundant ones.
\end{sloppypar}

\begin{sloppypar}
A second point about the visual stream deserves emphasis, because it is easily misread. It might be appealing to treat \condtrue{} as a control or a baseline against which the two distractor conditions are compared, on the grounds that when the image agrees with the world-knowledge answer no entrainment can occur. This framing is incorrect and we do not adopt it. The world-knowledge answer is known independently of the image, and constitutes the reference against which entrainment is measured; the \condtrue{} \emph{image} is not a baseline in that sense. Instead, all three visual conditions are potential entrainment conditions, differing in what feature of the image could plausibly do the pulling. In \condtrue{}, the image depicts the answer, but a model may still fixate on some other aspect of the image (a co-occurring object, a background element) and be pulled toward it. In \condreldist{}, an entity partially satisfying the query's semantic descriptors occupies the depicted content and can compete with the answer. In \condrandist{}, an entity that does not satisfy the query's semantic descriptors occupies the image, providing a completely different test of whether image content unrelated to the query nonetheless exerts a pull. Whether and how these three conditions produce different amounts of entrainment is exactly what the instrument is built to let one investigate.
\end{sloppypar}

\subsection{Design Invariants}
\label{sec:design}

Three statements per condition multiplies the number of effective probes and permits estimation of entrainment at the item level, through within-item variance across the three statements, rather than only at the corpus level. Every short-form distractor is constructed to differ from the item's answer: if a short trigger coincided with the correct answer, a model going with the trigger would be indistinguishable from a model answering correctly, so by construction every short trigger is a wrong answer and any pull toward it is unambiguously entrainment. For the visual stream, curation additionally targets two confound-reducing goals: distractor entities are chosen, to share no subword with the answer (for example, ``mango'' rather than ``pineapple'' as a distractor for ``apple,'' so that a measured pull reflects visual-semantic influence rather than surface-form overlap), and the relatable images within an item are chosen, wherever possible, with a similar tonal register, so that differences in behavior are less likely to be driven by low-level image statistics than by depicted content.

\subsection{Relationship to Prior Work}

ENTRAP-VL builds on the entrainment framing of \citet{niu2025llama} and departs from it in three ways. In \emph{modality}, prior work studies a single textual channel (in LLMs); we treat entrainment as dual (in VLMs). In \emph{construction}, prior work uses templatic prompts; we use manual, item-aware curation. In \emph{taxonomy}, prior work uses four context types; we use eight for textual entrainment. Table~\ref{tab:mapping} gives the correspondence.

\begin{table*}[t]
\caption{Correspondence between the four context types of \citet{niu2025llama} and the eight conditions of ENTRAP-VL with respect to the textual entrainment. Four conditions correspond directly or closely; four are novel. The counterfactual row is ``closest'' rather than ``direct'' because prior work attaches the falsehood to a semantically related neighbor entity of the queried subject, whereas ENTRAP-VL attaches it to the depicted scene itself.}
\vspace{0.6em}
\centering
\begin{tabular}{lll}
\toprule
\citet{niu2025llama} & ENTRAP-VL & Relationship \\
\midrule
Related (competing related entity) & \condreldist{} & direct \\
Irrelevant (unrelated full statement) & \condrandist{} & direct \\
Random (bare unrelated word) & \condranshort{} & direct \\
Counterfactual (on a related entity) & \condrelcf{} & closest; anchored to the depicted scene \\
\midrule
--- & \condtrue{} & novel: true context consistent with the scene \\
--- & \condcontra{} & novel: scene-relative falsehood \\
--- & \condrelshort{} & novel: bare scene-associated trigger \\
--- & \condrancf{} & novel: world-level falsehood, unassociated entity \\
\bottomrule
\end{tabular}
\label{tab:mapping}
\end{table*}

\section{ENTRAP-VL}
\label{sec:dataset}
ENTRAP-VL comprises two mirror streams across eight categories: \texttt{creatures}, \texttt{electronics}, \texttt{fruits}, \texttt{household\_objects}, \texttt{humans}, \texttt{monuments}, \texttt{natural\_landscapes}, and \texttt{vehicles}. Category labels denote the principal subject of the item and are organizational; as Section~\ref{sec:taxonomy} showed, a category label constrains neither the query type nor the relatability relation.

\begin{sloppypar}
\paragraph{\textbf{Textual stream (800 items).}} Each item (image) pairs with a textual query, a long- and short-form ground-truth answer, and the eight context conditions, each a list of three statements. Here, image acts as a perceptual referent. The image and textual query are fixed and the textual context is the accompanying variable. The stream contains all eight categories at 100 items each. The three-statements-per-condition design serves two purposes. It multiplies the number of effective probes: at the statement level the textual stream contains $800 \times 8 \times 3 = 19{,}200$ context-injections, so a full protocol produces far more measurements than the item count suggests. And it supports estimation of entrainment at the item level, through within-item variance across the three statements per condition, rather than only at the corpus level. This last point matters because the phenomenon of interest is a per-item susceptibility to context, and having multiple statements per condition per item is the minimum design that lets one estimate it.
\end{sloppypar}

The items span a range of query types. Activity recognition (``what is the dog doing?''), attribute identification (``what color is the bed?''), text reading in the image (``what does the mat say?''), detection of an incidental scene element (``what flying creatures are visible above the Taj Mahal?''), counting, etc., covering a variety but is not exhaustive. This diversity is a property of manual curation rather than an axis of the taxonomy: the taxonomy classifies context conditions, not query types. But query diversity matters for the interpretation of entrainment: pull may plausibly differ across query types, and having a stream that covers several allows for post-hoc analyses of this sort. Categories are organizational and, as illustrated in Section~\ref{sec:taxonomy}, do not constrain the query type or the relatability relation.

\begin{sloppypar}
\paragraph{\textbf{Visual stream (700 items).}} Each item (textual query), whose answer is world knowledge, pairs with three candidate images, one each for \condtrue{}, \condreldist{}, and \condrandist{}. The textual query is fixed and the image is the accompanying variable. The \texttt{humans} category is excluded from this stream by design (Section~\ref{sec:ethics}), leaving seven categories at 100 items each. As Section~\ref{sec:visual-taxonomy} argued, the three image conditions instantiate the association axis on truth-holding images, and all three are candidate entrainment conditions, not just the two distractors.
\end{sloppypar}

\begin{figure*}[t]
\centering
\setlength{\tabcolsep}{2pt}
\renewcommand{\arraystretch}{1.1}
\begin{tabular}{@{}l c c c@{}}
 & \condtrue{} & \condreldist{} & \condrandist{} \\
\rotatebox{90}{\parbox{2.6cm}{\centering \emph{creatures}\\ ``large gray animal''\\ $\to$ \textbf{elephant}}}
 & \includegraphics[width=0.28\textwidth]{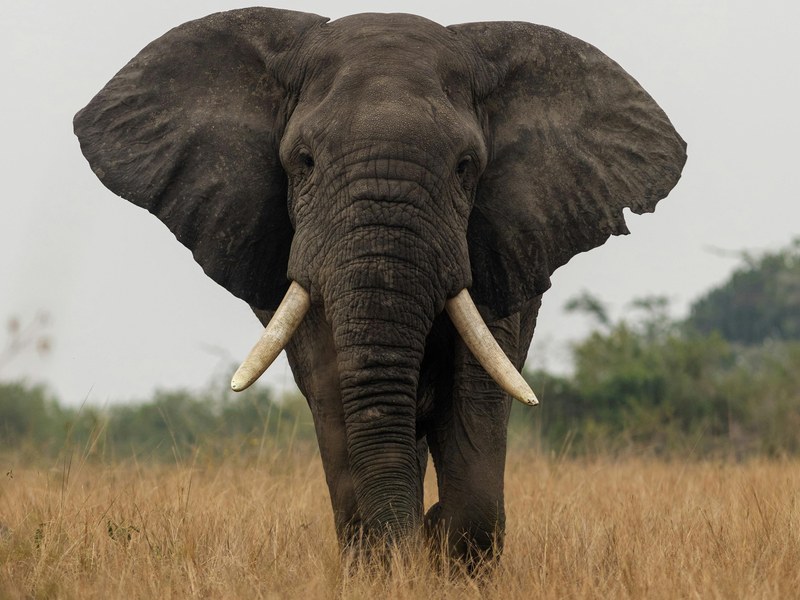}
 & \includegraphics[width=0.28\textwidth]{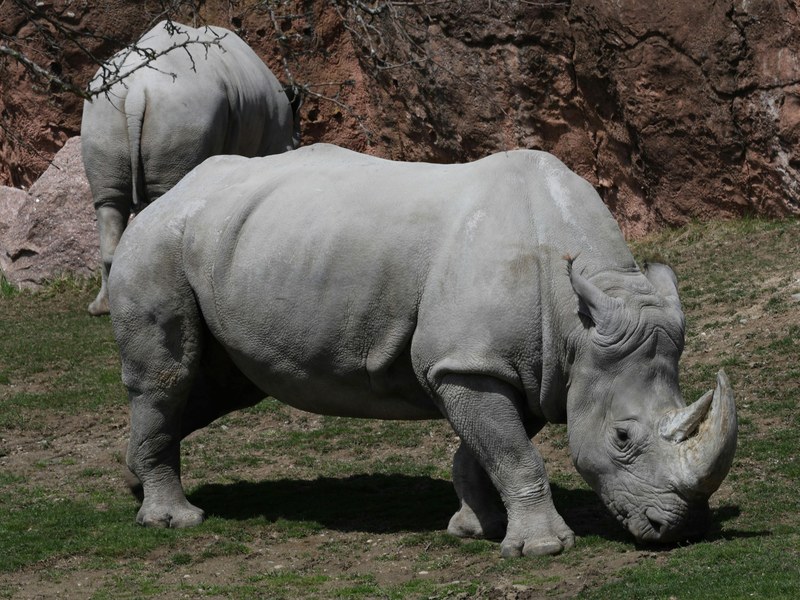}
 & \includegraphics[width=0.28\textwidth]{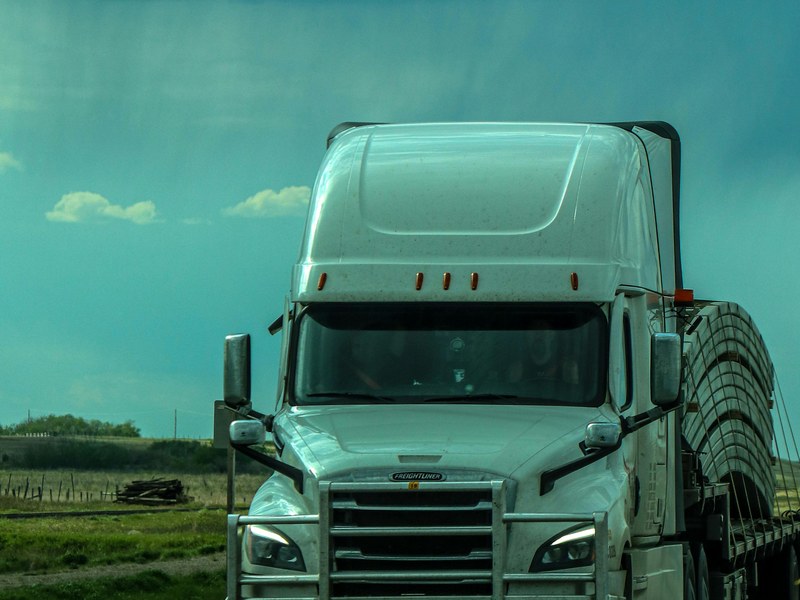} \\
\rotatebox{90}{\parbox{2.6cm}{\centering \emph{fruits}\\ ``keeps the doctor away''\\ $\to$ \textbf{apple}}}
 & \includegraphics[width=0.28\textwidth]{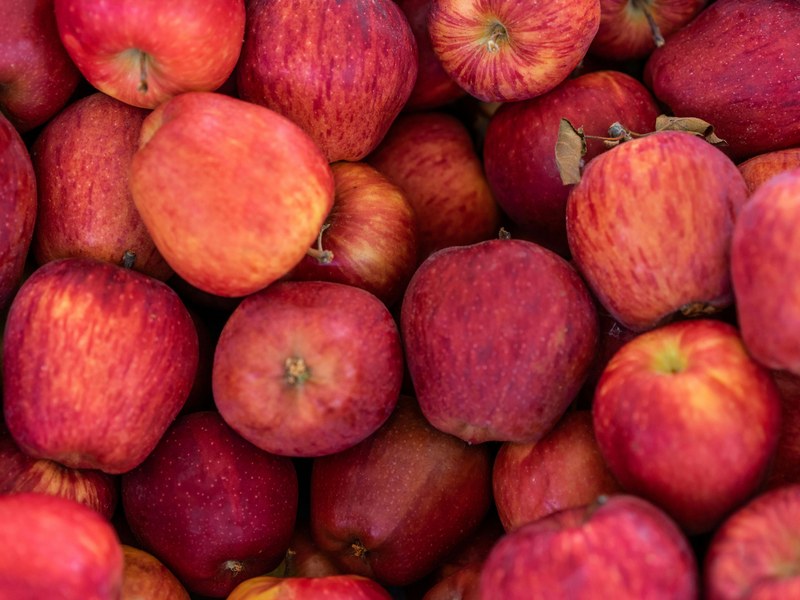}
 & \includegraphics[width=0.28\textwidth]{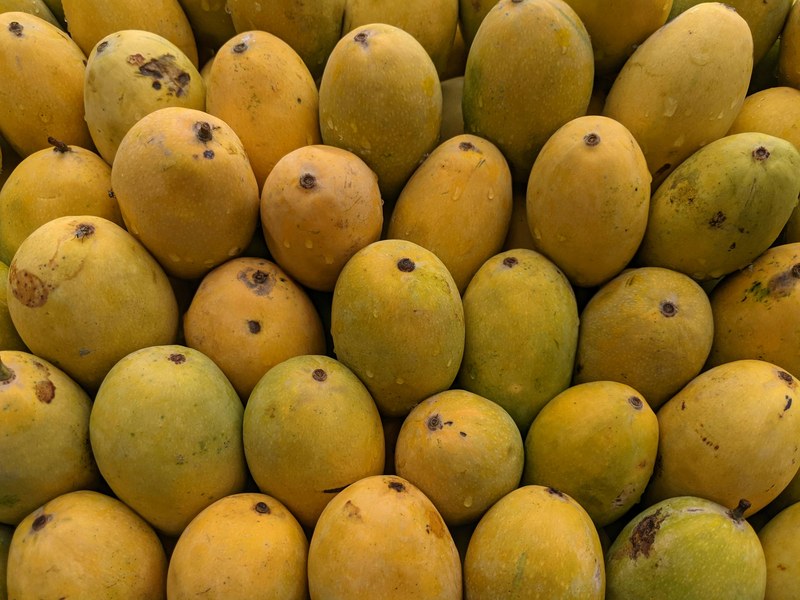}
 & \includegraphics[width=0.28\textwidth]{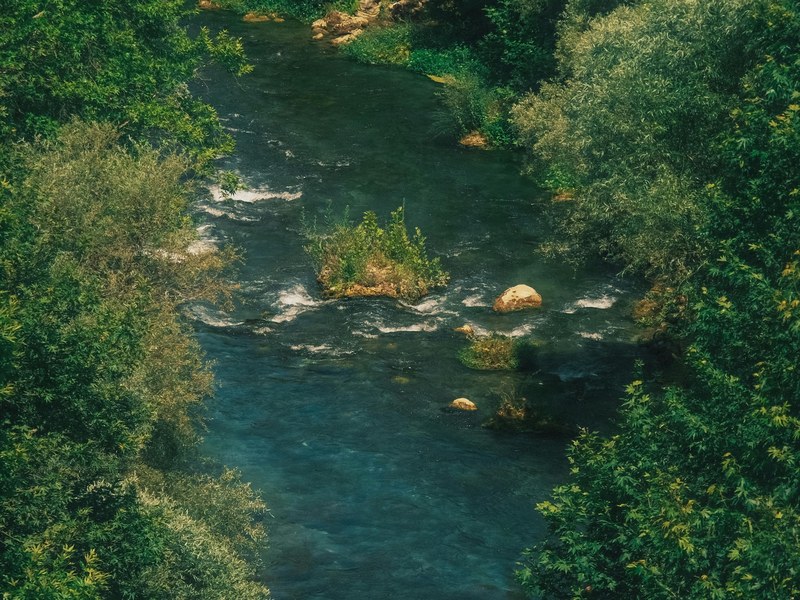} \\
\rotatebox{90}{\parbox{2.6cm}{\centering \emph{vehicles}\\ ``family drives on road''\\ $\to$ \textbf{car}}}
 & \includegraphics[width=0.28\textwidth]{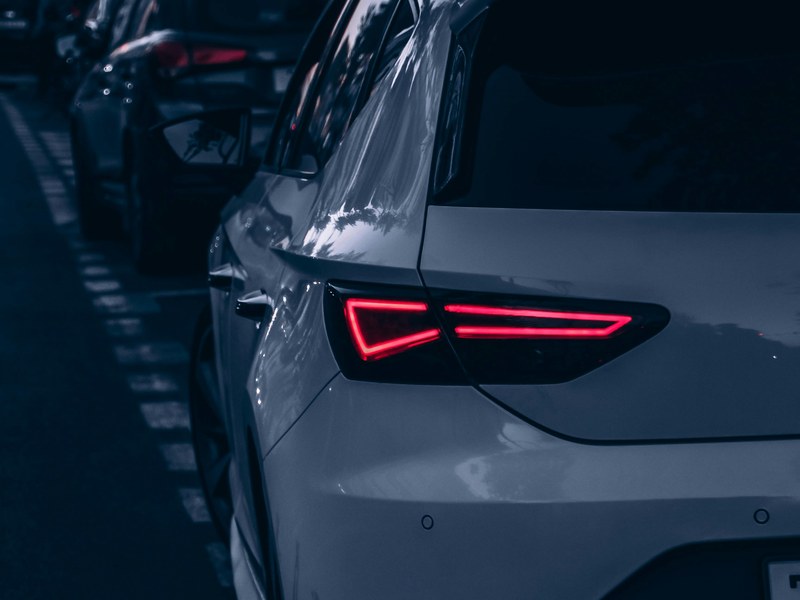}
 & \includegraphics[width=0.28\textwidth]{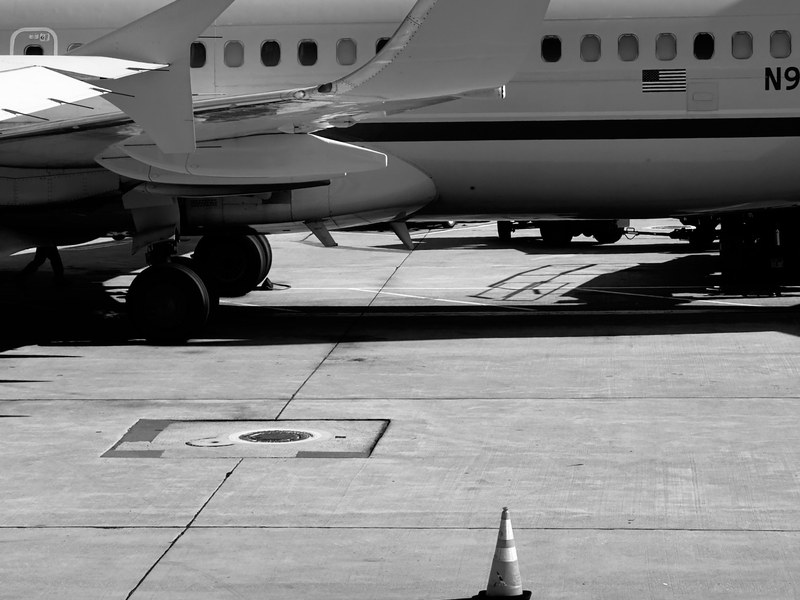}
 & \includegraphics[width=0.28\textwidth]{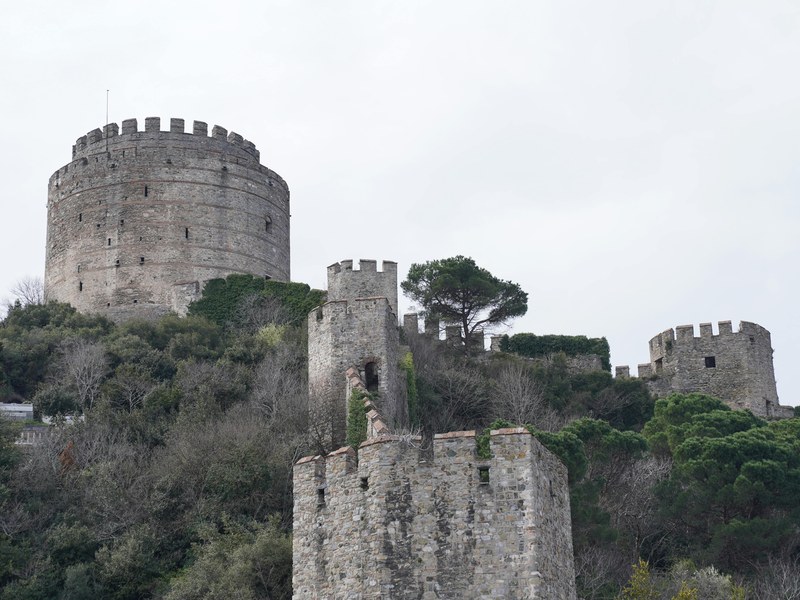} \\
\end{tabular}
\caption{Three items from the visual stream, one per row, showing the three image conditions. The abbreviated textual query and the ground-truth answer are shown at the left of each row depicting a unique category. In each row the \texttt{relatable\_true} image depicts the answer to the query, the \texttt{relatable\_distractor} depicts an entity partially satisfying the query's semantic descriptors while distinct from the answer (rhino for elephant, mango for apple, airplane for car), and the \texttt{random\_distractor} depicts an entity that does not satisfy the query's semantic descriptors. The vehicles row illustrates that relatability tracks satisfaction of the query's descriptors rather than visual similarity: a stone fortress is visually as imposing as a car or an airplane, yet is the random distractor because it doesn't satisfies the query's descriptors.}
\label{fig:visual-examples}
\end{figure*}

Three items illustrate the construction as shown in Figure~\ref{fig:visual-examples}. (1.) For ``A large gray animal with a trunk and tusks is an'' (answer \emph{elephant}), the \condtrue{} image shows an elephant, the \condreldist{} image shows a rhinoceros (satisfying \emph{large} and \emph{gray} while lacking trunk and tusks, hence a competing candidate rather than merely a same-category one; a goat, though categorically an animal, would not qualify because it satisfies neither of those descriptors), and the \condrandist{} image shows a truck. (2.) For ``A round red fruit that keeps the doctor away is an'' (answer \emph{apple}), the images are apples, mangoes, and a forest river respectively; the relatable distractor is a mango rather than a pineapple precisely to avoid sharing a subword with the answer (see Section~\ref{sec:design}). (3.) For ``A four-wheeled vehicle that a family drives on the road is a'' (answer \emph{car}), the images are a car, an airplane, and a fortress; the airplane is the relatable distractor because it shares the functional category of vehicle and can carry a family; the two images further show both vehicles on road having wheels, while the visually large fortress is random because it does not, a further instance of relatability cross-cutting visual similarity.

Two visual-stream design constraints are worth noting. First, all images were sourced from royalty-free repositories and verified free of watermarks prior to release: the presence of a watermark on a distractor image would introduce a per-image confound the taxonomy is not designed to control for. Second, per-item image selection targets two curation goals (Section~\ref{sec:design}): lexical disjointness of distractor entity names from the answer, and tonal matching within the item's three images. The first ensures that measured pull reflects visual-semantic influence rather than surface-form token overlap; the second reduces the risk that entrainment differences across conditions are driven by low-level image statistics rather than depicted content. These goals are pursued under image-availability constraints.

\paragraph{\textbf{Construction and provenance.}} The dataset was curated by the authors. The textual stream was authored manually, without generative assistance. For the visual stream, query prompts were generated with the assistance of a large language model and then reviewed and curated by the authors, with images sourced manually; this is acceptable for the visual stream because its prompts are simple factual elicitations whose phrasing carries little weight, while the entrainment signal is carried by the images. Images were sourced from royalty-free repositories (Unsplash, Pexels, Pixabay) and verified to be free of watermarks prior to release. Full schema, taxonomy, and provenance documentation will accompany the release, including a data statement.

\section{What the Instrument Enables}
\label{sec:protocols}

ENTRAP-VL is designed to support, but does not itself report, measurement of entrainment. We sketch the protocols it enables. In the textual stream, one feeds the model the fixed image and textual query together with each injected context condition and compares its answer to the answer it gives with no injected context and to the ground truth; a shift toward the content of the injected context is the Textual Entrainment signal. In the visual stream, one holds the textual query fixed and supplies each candidate image in turn. Because the answer is world knowledge that does not depend on the image, the reference against which entrainment is measured is the model's no-image answer to the textual query; any deviation from that reference under any of the three image conditions is the Visual Entrainment signal for that condition. As Section~\ref{sec:visual-taxonomy} makes explicit, all three visual conditions are candidate entrainment conditions, not just the two distractors: \condtrue{} can pull if the model fixates on some feature of the image other than the answer itself, and \condreldist{} and \condrandist{} can primarily pull toward the competing entity depicted in the image. What differs across the three is the type of image content available to pull the model.

\paragraph{\textbf{Measurements per item and per condition.}} At the most granular level, a protocol yields, for each item and each condition, the model's output distribution over some vocabulary or answer set. A range of derived measurements can then be computed. Among the most natural: the probability the model assigns to the injected entity or statement content, the probability it assigns to the ground-truth answer, and the change in either quantity relative to the no-context (or no-image) reference. The ratio or difference between the probability of the injected content and the probability of the ground-truth answer under each condition operationalizes the ``pull strength'' concept in Section~\ref{sec:position}. Because each textual condition has three statements, per-item variance across those three provides an internal reliability estimate for the measurement; because all seven visual categories share the same three-condition structure, per-condition measurements are commensurable across items and across categories.

\begin{sloppypar}
\paragraph{\textbf{Cross-condition comparisons the taxonomy makes possible.}} The taxonomy is what enables finer comparisons than any single condition permits. Within the textual stream, one can ask whether pull differs by veracity (does \condcontra{} context, false of the scene but plausible, pull more or less than \condrelcf{} context that is impossible?); by association (\condreldist{} vs.\ \condrandist{}, and \condrelshort{} vs.\ \condranshort{}); by form (full sentence vs.\ bare entity trigger); and by combinations thereof. The design invariant that short-form triggers are disjoint from the answer makes the short-form pull cleanly attributable: any pull toward the trigger is unambiguously entrainment rather than correctness. Within the visual stream, the relatable/random contrast across images (\condreldist{} vs.\ \condrandist{}) is the direct analog of the short-form contrast in the textual stream, since both hold the trigger content fixed as an entity and vary only whether it satisfies the query's semantic descriptors. It is also interpretable as a full-form contrast because the images contain not only the specific trigger content but also the surroundings.
\end{sloppypar}

\paragraph{\textbf{The central cross-stream comparison.}} The dual structure supports the comparison for which the instrument was principally built: within a single model, whether the model is entrained by textual context and visual context to comparable degrees. This is the empirical version of the dual-modality argument in Section~\ref{sec:position}. It has no direct precedent in prior work, which studies each channel in isolation, and no direct proxy in existing distraction benchmarks, which do not separate the two sources of pull by construction. The two streams differ in the number and type of conditions they instantiate, but the shared association axis and the shared world-knowledge-answer framework provide a common footing for analogous quantities to be computed and juxtaposed. We state this as a question the instrument opens, not a finding.

\section{Limitations and Ethics}
\label{sec:ethics}

\paragraph{\textbf{Scale and locale.}} At 1{,}500 items, the dataset is a focused diagnostic, not a large-scale benchmark, and is English-only. The three-statements-per-condition design mitigates the size at the item level but does not substitute for scale. Category and item selection reflect the curators' choices and image availability and may carry corresponding inherent selection intuitions. Results obtained on ENTRAP-VL should be read as evidence about the specific conditions the dataset instantiates, not as population-level estimates of entrainment across all possible queries, images, or contexts. Multilingual extensions and larger-scale variants are natural directions the taxonomy would support in principle, and we regard them as valuable follow-ups.

\paragraph{\textbf{Curation subjectivity and validation.}} Judgments of relatability, of local versus global falsehood, and of clean distractors are made by co-authors and reviewed by the leads. The final to-be-released artifact is checked for structural invariants, including schema completeness, image resolution, watermarking, identifier uniqueness, and disjointness of short-form triggers from the answer. The visual-stream confound-reduction goals of Section~\ref{sec:design}, i.e., lexical disjointness of distractor entity names from the answer and tonal matching within an item, are pursued under image-availability constraints. Users needing strict control on either axis should verify it on the specific items or categories they use. The taxonomy specification itself will be provided as documentation with the release, so that alternative curators can extend or replicate the dataset with the same design principles.

\paragraph{\textbf{People in the dataset.}} The \texttt{humans} category contains images of real people from royalty-free repositories. Every query about a person is answered by something directly observable: an action; a visible physical state; or an occupation indicated by visible cues, with no person identification and no inference of protected or internal attributes. Where a query uses affect-adjacent wording, the answer is the observable action rather than an inferred feeling (e.g., ``screaming,'' not an inferred emotional state). The category is excluded entirely from the visual stream, since constructing visual distractors for human subjects would require pairing additional images of identifiable people, a consent and identification concern the dataset does not incur.

\paragraph{\textbf{Responsible use.}} ENTRAP-VL is a diagnostic instrument, not a leaderboard or accuracy benchmark. The taxonomy specifies context conditions for which pull is anticipated, not tasks a good model should always answer ``correctly.'' A model that exhibits reduced pull under distractor conditions is behaving well by the criterion the instrument measures; a model that exhibits large pull is exhibiting the phenomenon the instrument is designed to detect. Neither is a general endorsement or condemnation of the model. Downstream users are encouraged to report results by condition rather than aggregated, and to include the no-context (or no-image) reference measurements alongside condition-specific measurements, so that entrainment magnitudes can be interpreted meaningfully.

\section{Conclusion}

We have taken the position that contextual entrainment in vision-language models is a distinct, dual phenomenon whose study requires a purpose-built, taxonomically structured instrument. The move to VLMs makes entrainment drivable independently through visual and textual context, and opens a scene-relative veracity distinction with no counterpart in the text-only formulation. ENTRAP-VL embodies this position as a usable instrument: a manually curated, dual-stream, taxonomically organized dataset with the documentation and protocols needed to probe entrainment rigorously. We will release it so that the open question of whether and how VLMs are entrained by context can be taken up by the community.

\newpage
\bibliographystyle{ACM-Reference-Format}
\bibliography{entrap_vl}

@inproceedings{niu2025llama,
  title={Llama see, llama do: A mechanistic perspective on contextual entrainment and distraction in LLMs},
  author={Niu, Jingcheng and Yuan, Xingdi and Wang, Tong and Saghir, Hamidreza and Abdi, Amir H},
  booktitle={Proceedings of the 63rd Annual Meeting of the Association for Computational Linguistics (Volume 1: Long Papers)},
  pages={16218--16239},
  year={2025}
}

@inproceedings{shi2023large,
  title={Large language models can be easily distracted by irrelevant context},
  author={Shi, Freda and Chen, Xinyun and Misra, Kanishka and Scales, Nathan and Dohan, David and Chi, Ed H and Sch{\"a}rli, Nathanael and Zhou, Denny},
  booktitle={International Conference on Machine Learning},
  pages={31210--31227},
  year={2023},
  organization={PMLR}
}

@inproceedings{sharma2024losing,
  title={Losing visual needles in image haystacks: Vision language models are easily distracted in short and long contexts},
  author={Sharma, Aditya and Saxon, Michael and Wang, William Yang},
  booktitle={Findings of the Association for Computational Linguistics: EMNLP 2024},
  pages={5429--5451},
  year={2024}
}

@article{golovanevsky2024vlms,
  title={What do vlms notice? a mechanistic interpretability pipeline for gaussian-noise-free text-image corruption and evaluation},
  author={Golovanevsky, Michal and Rudman, William and Palit, Vedant and Singh, Ritambhara and Eickhoff, Carsten},
  journal={arXiv preprint arXiv:2406.16320},
  year={2024}
}

@inproceedings{wang2025v,
  title={V-seam: Visual semantic editing and attention modulating for causal interpretability of vision-language models},
  author={Wang, Qidong and Hu, Junjie and Jiang, Ming},
  booktitle={Proceedings of the 2025 Conference on Empirical Methods in Natural Language Processing},
  pages={17407--17431},
  year={2025}
}

@article{olsson2022context,
  title={In-context learning and induction heads},
  author={Olsson, Catherine and Elhage, Nelson and Nanda, Neel and Joseph, Nicholas and DasSarma, Nova and Henighan, Tom and Mann, Ben and Askell, Amanda and Bai, Yuntao and Chen, Anna and others},
  journal={arXiv preprint arXiv:2209.11895},
  year={2022}
}

@inproceedings{xu2024knowledge,
  title={Knowledge conflicts for llms: A survey},
  author={Xu, Rongwu and Qi, Zehan and Guo, Zhijiang and Wang, Cunxiang and Wang, Hongru and Zhang, Yue and Xu, Wei},
  booktitle={Proceedings of the 2024 Conference on Empirical Methods in Natural Language Processing},
  pages={8541--8565},
  year={2024}
}

@inproceedings{longpre2021entity,
  title={Entity-based knowledge conflicts in question answering},
  author={Longpre, Shayne and Perisetla, Kartik and Chen, Anthony and Ramesh, Nikhil and DuBois, Chris and Singh, Sameer},
  booktitle={Proceedings of the 2021 conference on empirical methods in natural language processing},
  pages={7052--7063},
  year={2021}
}

@inproceedings{xie2024adaptive,
  title={Adaptive chameleon or stubborn sloth: Revealing the behavior of large language models in knowledge conflicts},
  author={Xie, Jian and Zhang, Kai and Chen, Jiangjie and Lou, Renze and Su, Yu},
  booktitle={International Conference on Learning Representations},
  volume={2024},
  pages={35623--35646},
  year={2024}
}

@inproceedings{shi2024trusting,
  title={Trusting your evidence: Hallucinate less with context-aware decoding},
  author={Shi, Weijia and Han, Xiaochuang and Lewis, Mike and Tsvetkov, Yulia and Zettlemoyer, Luke and Yih, Wen-tau},
  booktitle={Proceedings of the 2024 Conference of the North American Chapter of the Association for Computational Linguistics: Human Language Technologies (Volume 2: Short Papers)},
  pages={783--791},
  year={2024}
}

@inproceedings{du2024context,
  title={Context versus prior knowledge in language models},
  author={Du, Kevin and Sn{\ae}bjarnarson, V{\'e}steinn and Stoehr, Niklas and White, Jennifer and Schein, Aaron and Cotterell, Ryan},
  booktitle={Proceedings of the 62nd Annual Meeting of the Association for Computational Linguistics (Volume 1: Long Papers)},
  pages={13211--13235},
  year={2024}
}

@inproceedings{chen2024benchmarking,
  title={Benchmarking large language models in retrieval-augmented generation},
  author={Chen, Jiawei and Lin, Hongyu and Han, Xianpei and Sun, Le},
  booktitle={Proceedings of the AAAI Conference on Artificial Intelligence},
  volume={38},
  number={16},
  pages={17754--17762},
  year={2024}
}

@inproceedings{fang2024enhancing,
  title={Enhancing noise robustness of retrieval-augmented language models with adaptive adversarial training},
  author={Fang, Feiteng and Bai, Yuelin and Ni, Shiwen and Yang, Min and Chen, Xiaojun and Xu, Ruifeng},
  booktitle={Proceedings of the 62nd Annual Meeting of the Association for Computational Linguistics (Volume 1: Long Papers)},
  pages={10028--10039},
  year={2024}
}

@article{liu2024insight,
  title={Insight over sight: Exploring the vision-knowledge conflicts in multimodal llms},
  author={Liu, Xiaoyuan and Wang, Wenxuan and Yuan, Youliang and Huang, Jen-tse and Liu, Qiuzhi and He, Pinjia and Tu, Zhaopeng},
  journal={arXiv preprint arXiv:2410.08145},
  year={2024}
}

@article{zhu2024unraveling,
  title={Unraveling cross-modality knowledge conflicts in large vision-language models},
  author={Zhu, Tinghui and Liu, Qin and Wang, Fei and Tu, Zhengzhong and Chen, Muhao},
  journal={arXiv preprint arXiv:2410.03659},
  year={2024}
}

@inproceedings{jia2026benchmarking,
  title={Benchmarking multimodal knowledge conflict for large multimodal models},
  author={Jia, Yifan and Du, Yuntao and Jiang, Kailin and Liang, Yuyang and Ren, Qihan and Xin, Yi and Yang, Rui and Feng, Fenze and Chen, MingCai and Lu, Hengyang and others},
  booktitle={Proceedings of the AAAI Conference on Artificial Intelligence},
  volume={40},
  number={27},
  pages={22283--22291},
  year={2026}
}

@inproceedings{lee2025vlind,
  title={Vlind-bench: Measuring language priors in large vision-language models},
  author={Lee, Kang-il and Kim, Minbeom and Yoon, Seunghyun and Kim, Minsung and Lee, Dongryeol and Koh, Hyukhun and Jung, Kyomin},
  booktitle={Findings of the Association for Computational Linguistics: NAACL 2025},
  pages={4129--4144},
  year={2025}
}

@inproceedings{bitton2023breaking,
  title={Breaking common sense: Whoops! a vision-and-language benchmark of synthetic and compositional images},
  author={Bitton-Guetta, Nitzan and Bitton, Yonatan and Hessel, Jack and Schmidt, Ludwig and Elovici, Yuval and Stanovsky, Gabriel and Schwartz, Roy},
  booktitle={Proceedings of the IEEE/CVF International Conference on Computer Vision},
  pages={2616--2627},
  year={2023}
}

@article{goyal2026expense,
  title={The Expense of Seeing: Attaining Trustworthy Multimodal Reasoning Within the Monolithic Paradigm},
  author={Goyal, Karan},
  journal={arXiv preprint arXiv:2604.20665},
  year={2026}
}

@inproceedings{li2025have,
  title={Have the VLMs lost confidence? A study of sycophancy in VLMs},
  author={Li, Shuo and Ji, Tao and Fan, Xiaoran and Lu, Linsheng and Yang, Leyi and Yang, Yuming and Xi, Zhiheng and Zheng, Rui and Wang, Yuran and Gui, Tao and others},
  booktitle={International Conference on Learning Representations},
  volume={2025},
  pages={2739--2759},
  year={2025}
}

@inproceedings{li2023evaluating,
  title={Evaluating object hallucination in large vision-language models},
  author={Li, Yifan and Du, Yifan and Zhou, Kun and Wang, Jinpeng and Zhao, Xin and Wen, Ji-Rong},
  booktitle={Proceedings of the 2023 conference on empirical methods in natural language processing},
  pages={292--305},
  year={2023}
}

@inproceedings{rohrbach2018object,
  title={Object hallucination in image captioning},
  author={Rohrbach, Anna and Hendricks, Lisa Anne and Burns, Kaylee and Darrell, Trevor and Saenko, Kate},
  booktitle={Proceedings of the 2018 Conference on Empirical Methods in Natural Language Processing},
  pages={4035--4045},
  year={2018}
}

@article{liu2024lost,
  title={Lost in the middle: How language models use long contexts},
  author={Liu, Nelson F and Lin, Kevin and Hewitt, John and Paranjape, Ashwin and Bevilacqua, Michele and Petroni, Fabio and Liang, Percy},
  journal={Transactions of the association for computational linguistics},
  volume={12},
  pages={157--173},
  year={2024}
}

@inproceedings{goyal2017making,
  title={Making the v in vqa matter: Elevating the role of image understanding in visual question answering},
  author={Goyal, Yash and Khot, Tejas and Summers-Stay, Douglas and Batra, Dhruv and Parikh, Devi},
  booktitle={Proceedings of the IEEE conference on computer vision and pattern recognition},
  pages={6904--6913},
  year={2017}
}

\end{document}